\documentclass[letterpaper, 10 pt, conference]{ieeeconf}  

\IEEEoverridecommandlockouts                              

\usepackage{cite}
\usepackage{amsmath,amssymb,amsfonts}
\usepackage{algorithmic}
\usepackage{graphicx}
\usepackage{textcomp}
\usepackage{multirow}
\usepackage{cuted}
\usepackage{capt-of}
\usepackage{float}
\usepackage{booktabs}
\usepackage{threeparttable}
\usepackage{mathtools}
\usepackage[mathscr]{eucal}
\usepackage[table,xcdraw]{xcolor}
\usepackage{colortbl}
\usepackage{colortbl}

\usepackage{balance}
\usepackage{color}

\usepackage{marvosym}
\usepackage{algorithmic}
\usepackage{indentfirst}
\usepackage{makecell}

\makeatletter
\let\NAT@parse\undefined
\makeatother
\usepackage{hyperref}

\title{\LARGE \bf
UncAD: Towards Safe End-to-end Autonomous Driving via Online Map Uncertainty 
}

\author{Pengxuan Yang$^{1,2,3*}$, Yupeng Zheng$^{2,3*}$, Qichao Zhang$^{2,3\dagger}$, Kefei Zhu$^{2}$, Zebin Xing$^{2,3}$, \\ Qiao Lin$^{4}$,  Yun-Fu Liu$^{4}$, Zhiguo Su$^{4}$, Dongbin Zhao$^{2,3}$ 
\thanks{$^{1}$Key Laboratory of Safety Intelligent Mining in Non-coal Open-pit Mines, National Mine safety Administration, Guangdong Guangzhou, 510000, China}
\thanks{$^{2}$The State Key Laboratory of Multimodal Artificial Intelligence Systems, Institute of Automation, Chinese Academy of Sciences}
\thanks{$^{3}$School of Artificial Intelligence, University of Chinese Academy of Sciences, Beijing, China,}%
\thanks{$^{4}$EACON, Fujian, China}
\thanks{$*$ Equal contribution.}
\thanks{$\dagger$ Corresponding author.}
\thanks{This work is supported by the National Key Research and Development Program of China under Grants 2022YFA1004000, the Beijing Natural Science Foundation under No. 4242052, and the State Key Laboratory of Safety Intelligent Mining in Non-coal Open-pit Mines, National Mine Safety Administration (Grant No.2024-ZD04).}
}

\begin{document}

\maketitle
\thispagestyle{empty}
\pagestyle{empty}


\begin{abstract}

End-to-end autonomous driving aims to produce planning trajectories from raw sensors directly.
Currently, most approaches integrate perception, prediction, and planning modules into a fully differentiable network, promising great scalability. 
However, these methods typically rely on deterministic modeling of online maps in the perception module for guiding or constraining vehicle planning, which may incorporate erroneous perception information and further compromise planning safety.
To address this issue, we delve into the importance of online map uncertainty for enhancing autonomous driving safety and propose a novel paradigm named UncAD.
Specifically, UncAD first estimates the uncertainty of the online map in the perception module. 
It then leverages the uncertainty to guide motion prediction and planning modules to produce multi-modal trajectories. 
Finally, to achieve safer autonomous driving, UncAD proposes an uncertainty-collision-aware planning selection strategy according to the online map uncertainty to evaluate and select the best trajectory.
In this study, we incorporate UncAD into various state-of-the-art (SOTA) end-to-end methods. 
Experiments on the nuScenes dataset show that integrating UncAD, with only a 1.9\% increase in parameters, can reduce collision rates by up to 26\% and drivable area conflict rate by up to 42\%.
Codes, pre-trained models, and demo videos can be accessed at \href{https://github.com/pengxuanyang/UncAD}{https://github.com/pengxuanyang/UncAD}.

\end{abstract}

\section{Introduction}
Recently, the end-to-end autonomous driving paradigm is treated as the most critical technology for autonomous driving and attracts increasing attention due to the strong scalable capability and application potential\cite{chen2024end, chib2023recent,zheng2024preliminary}.  

\begin{figure}[t]
    \centering
    \includegraphics[width=0.48\textwidth]{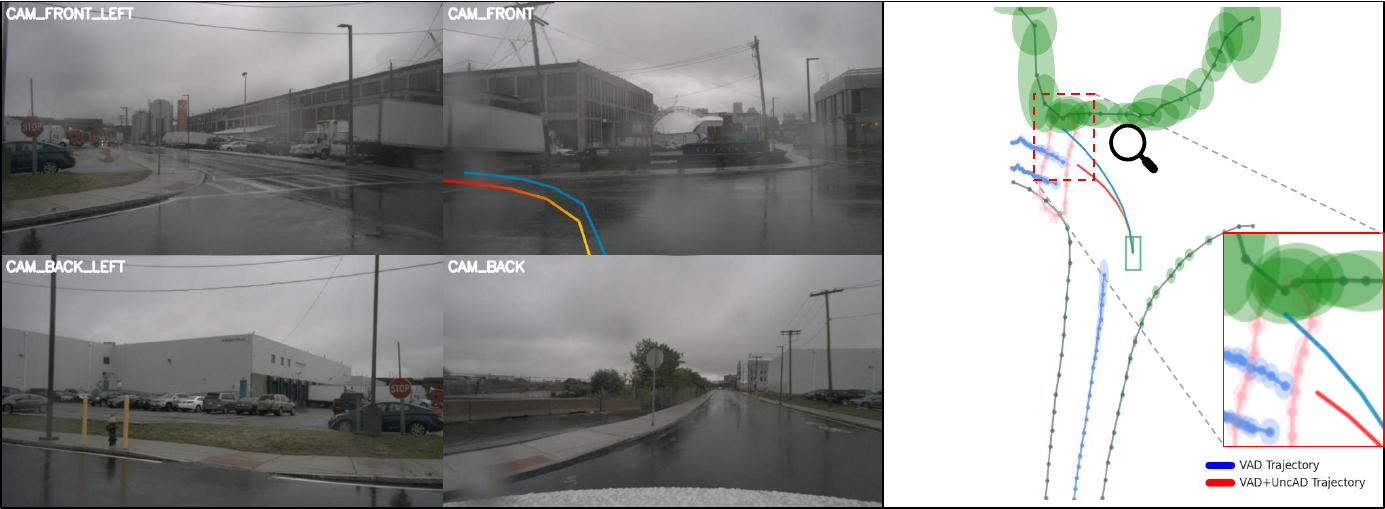} 
    \caption{The panoptic images are shown on the left, and the bird's eye view (BEV) map with the ego vehicle’s planning trajectory is shown on the right. 
    In the BEV map, green, pink, and blue ellipses represent the uncertainty of boundaries, pedestrians, and lane dividers, respectively. 
    In this challenging (rainy and sharp turning) scenario, VAD suffers from the inaccuracy online map, leading to collide with map boundaries (the blue trajectory in both sides). 
    Our method leverages map uncertainty to guide the ego vehicle to avoid high-uncertainty areas and drive within confirmed drivable area, resulting in safer trajectory (the red trajectory in both sides). 
    }
    \label{teaser}
    \vspace{-1.5mm}
\end{figure}

State-of-the-art methods have made great efforts on representation\cite{jiang2023vad, sun2024sparsedrive, zheng2024genad, chen2024vadv2,li2023planning}, multi-modal fusion\cite{chen2024dualat, ye2023fusionad}, and task design\cite{hu2022st, hu2023planning} to advance end-to-end autonomous driving.
However, most of them share the same limitation: 
\textit{Inability to estimate the online map uncertainty}.
They depend on deterministic modeling of online maps, which are highly sensitive to map errors and misalignments.
Such vulnerabilities can lead to misguided directions or constraints, resulting in unsafe planning, especially in sharp turn scenarios demonstrated in Fig. \ref{teaser} and Tab. \ref{tab2}.
The end-to-end paradigm should consider such perception uncertainty for safe deployment in real world.

To address this critical issue, we propose \textbf{UncAD}, a novel paradigm that integrates online map uncertainty into end-to-end autonomous driving systems as a guidance to produce, evaluate, and select safer and more reliable planning trajectories.
Specifically, we (1) introduce a Map Uncertainty Estimation (MUE) module that estimates map uncertainty by predicting map coordinates and its Laplace distribution.
Then, we (2) propose an Uncertainty-Guided Prediction and Planning (UGPnP) module, which allows the downstream prediction and planning to take the map uncertainty into account. It leverages the uncertainty modeled by the MUE module as a guidance to predict other agents' motion and produce multi-modal ego-centric planning trajectories.
To ensure the safety and reliability of planning trajectories, we (3) design an Uncertainty-Collision-Aware Selection (UCAS) strategy. It leverages motion prediction of other agents and map uncertainty to evaluate the planning multi-modal trajectories and filter out those at risk of collisions or traversing high-uncertainty areas.

To comprehensively explore the end-to-end planning under the map uncertainty, we integrated our method into two SOTA end-to-end autonomous methods, VAD\cite{jiang2023vad} with vectorized scene representation and SparseDrive\cite{sun2024sparsedrive} with sparse scene representation. 
Furthermore, we introduce a new metric \textit{Drivable Area Conflict Rate} (DACR) to enhance the comprehensive assessment of planning safety, addressing the shortcomings of commonly used metrics Displacement Error (DE) and Collision Rate (CR).
Leveraging our designed approach, UncAD achieves state-of-the-art performance on the nuScenes dataset \cite{caesar2020nuscenes}, demonstrating the effectiveness of map uncertainty estimation in end-to-end planning.

Our key contributions can be summarized as follows:
\begin{enumerate}
    \item We propose a novel paradigm termed UncAD, which integrates online map uncertainty into end-to-end autonomous driving systems.
    \item We design an Uncertainty-Guided Prediction and Planning module and an Uncertainty-Collision-Aware Selection strategy to comprehensively incorporate map uncertainty for safer planning. 
    \item We introduce a new metric Drivable Area Conflict Rate (DACR) for comprehensive assess the planning quality.
    \item Our method achieves SOTA performance on the nuScenes dataset under a comprehensive assessment. 
\end{enumerate}

\begin{table}[ht]
    \centering
    \setlength{\tabcolsep}{0.020\linewidth}
    \renewcommand{\arraystretch}{1.2} 
    \caption{Planning performance of VAD in Different Scenarios}
    \resizebox{0.35\textwidth}{!}{
    \begin{tabular}{cccc}
        \toprule  
        Scenario & L2(m) ↓ & Collision(\%) ↓ \\ 
        \midrule  
        Turn & \textbf{0.87} &  \textbf{0.29} \\ 
        Straight & 0.70 &  0.21 \\ 
        Overall & 0.72 &  0.23 \\ 
        \bottomrule  
    \end{tabular}
    \label{tab2}
    }
    \vspace{-2mm}
\end{table}

\section{Related Works}

\subsection{End to End Planning}

End-to-end planning has been a long goal for autonomous driving. 
As pioneers in end-to-end driving, ST-P3\cite{hu2022st} introduced an end-to-end pipeline that integrates perception, prediction, and planning. 
UniAD\cite{hu2023planning} advances it with innovative unified query design to incorporate multiple tasks into a differentiable network with a focus on optimizing ego-centric planning. 
Some studies \cite{jiang2023vad, sun2024sparsedrive} further refined scene representation, while others\cite{ye2023fusionad, zhang2022mmfn, chen2024dualat} boosted performance by delving into the fusion strategy of multi-modal information. 
Furthermore, more efforts such as incorporating generative modeling of motion prediction\cite{zheng2024genad,li2023conditional,li2023planning}, jointly learning prediction and planning\cite{fu2023interactionnet}, and utilizing multi-modal foundational models\cite{xu2024drivegpt4,jin2024tod3cap,jin2023adapt} or reinforcement learning techniques \cite{liu2024deep,wang2020dynamic,wang2024prototypical} have been made to improve capabilities of interaction, generalization, interpretability for autonomous driving systems.
However, most of these works ignore the importance of perception uncertainty, especially the online map uncertainty. 
Grounded in SOTA methods VAD\cite{jiang2023vad} and Sparsedrive\cite{sun2024sparsedrive}, our research focuses on the overlooked aspect, incorporating map uncertainty to achieve safe and reliable end-to-end autonomous driving in real world.

\subsection{Uncertainty in Deep Learning and Robotics}
Uncertainty estimation which predicts the confidence in model predictions is a classic and important problem in the fields of deep learning and robotics. 
Uncertainty is typically categorized into epistemic uncertainty (model uncertainty) and aleatoric uncertainty (data uncertainty) \cite{kendall2017uncertainties}. 

Due to the potential of capturing the noise inherent in data and improving performance, recent research has increasingly explored uncertainty in the field of 2D scene perception, such as normal estimation\cite{bae2021estimating,long2024adaptive} and depth estimation \cite{eldesokey2020uncertainty,kim2019laf,poggi2020uncertainty, zheng2023steps,michelmore2020uncertainty}.
In the field of robotics planning, some studies employed uncertainty to enhance the robustness of long-horizon motion planning\cite{kurniawati2011motion}, visual odometry\cite{yang2020d3vo}, UAV motion planning\cite{pairet2018uncertainty} and general planning modeling\cite{loquercio2020general} in real world.
Besides, Rohan Sinha et al.\cite{sinha2024real} explored the application of Large Language Models (LLMs) to evaluate uncertainty as a switcher for slow-and-fast system.
Recently, Gu et al. \cite{gu2024producing} lifted uncertainty estimation from 2D perception into 3D BEV representation, modeling map uncertainty via Laplace distribution. 
Building on the development of uncertainty estimation and application, our work proposes the first uncertainty-aware end-to-end autonomous planning framework, aiming to bridge the perception uncertainty and the application in planning for robustness and safety.

\section{Method}\label{3}
\begin{figure*}[t]
    \centering
    \includegraphics[width=\textwidth]{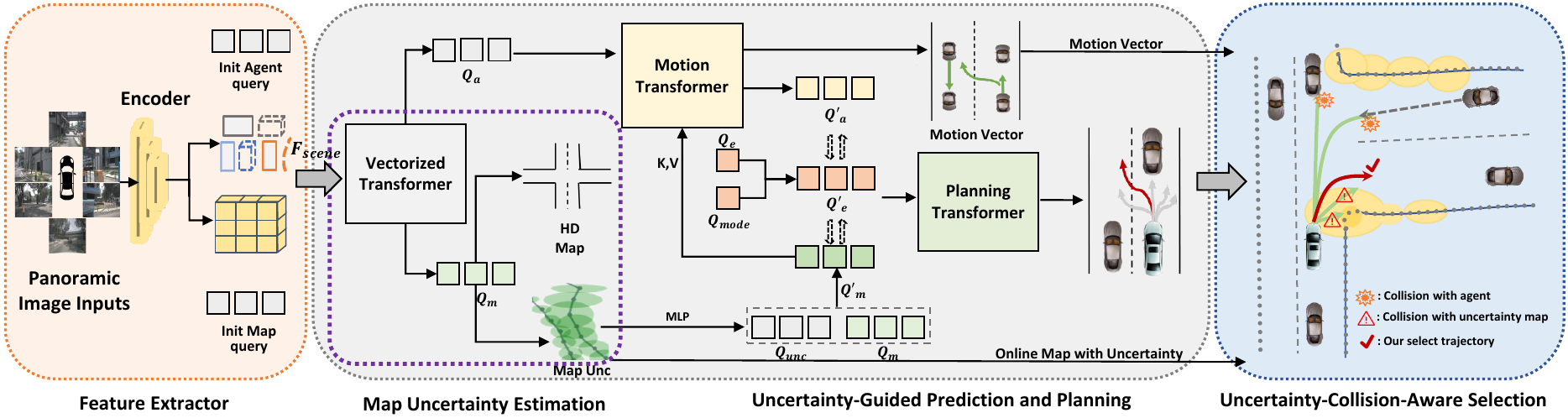} 
    \caption{The overall architecture of UncAD consists of four core modules. First, the Feature Extractor uses an encoder to project image inputs into BEV or sparse features. The Map Uncertainty Estimation (MUE) module encodes scene information into map queries and estimates map uncertainty. In the Uncertainty-Guided Prediction and Planning (UGPnP) module, agent and ego queries incorporate map uncertainty through query interaction, generating prediction and planning trajectories. Finally, the Uncertainty-Collision-Aware Selection (UCAS) strategy selects the optimal trajectory based on driving area uncertainty and collision risk, ensuring the safest path.
    }
    \label{mainfig}
    \vspace{-3mm}
\end{figure*}

The overall framework of UncAD is shown in Fig.\ref{mainfig}. 
In Section \ref{3A}, we provide an overview of the current end-to-end planning process. Section \ref{3B} introduces the Map Uncertainty Estimation (MUE) module, which estimates map uncertainty and integrates it into the perception module. In Section \ref{3C}, we present the Uncertainty-Guided Prediction and Planning (UGPnP) module, which incorporates map uncertainty to guide both prediction and multi-modal planning tasks. Finally, in Section \ref{3D}, we propose the Uncertainty-Collision-Aware Selection(UCAS) module, which selects the safest trajectory from the multi-modal trajectories by evaluating the driving area uncertainty and collision risks.

\subsection{Conventional End to End Planning}\label{3A}
Most end-to-end systems consist of three primary stages: perception, motion prediction, and planning. 
In the perception stage, the system encodes multi-view images into a BEV-centric \cite{chen2024vadv2,hu2023planning,zheng2024genad,jiang2023vad} or sparse-centric \cite{sun2024sparsedrive} intermediate representation \(F_{scene}\), which is typically used for deterministic online map construction and encoded as map features \(F_{map}\). 
In the motion prediction stage, the system combines \(F_{scene}\) and \(F_{map}\) to predict the intentions and trajectories of surrounding agents, producing agent features \(F_{agent}\). 
Finally, in the planning stage, the system integrates \(F_{scene}\), \(F_{map}\) and \(F_{agent}\) to learn the ego vehicle’s features \(F_{ego}\), and generates a safe planning trajectory \(\tau\). This process can be formalized as:
\begin{equation}
\begin{aligned}
    &F_{map} = \textbf{MapEncoder}(F_{scene}),\\
     &F_{agent}= \textbf{AgentEncoder}(F_{scene}, F_{map}),\\
    &F_{ego} = \textbf{EgoEncoder}(F_{scene}, F_{map}, F_{agent}),\\ 
    &\tau = \textbf{MLP}(F_{ego}).
\end{aligned}
\end{equation}

\subsection{Map Uncertainty Estimation}\label{3B}

Previous works \cite{jiang2023vad,liao2022maptr,liao2023maptrv2,liu2023vectormapnet,sun2024sparsedrive} on vectorized online maps employ a deterministic modeling for the map construction. 
In contrast, we introduce uncertainty into map element points to enrich online map information, facilitating safer trajectory planning. 

\textbf{Producing Online Map Uncertainty.}
Following Gu et al.\cite{gu2024producing}, we model each map element point $\mathbf{p} = (\boldsymbol{\mu},\mathbf{b})$ using a Laplace distribution, where $\boldsymbol{\mu}=(\mu_1,\mu_2)$ is a 2D vector representing normalized BEV $(x, y)$ coordinates and $\mathbf{b}$ denotes uncertainty parameters associated with the predicted point.
The joint probability density for a map element $M$ with $N$ points, denoted $M = \left\{ \mathbf{p}^{(i)} \right\}_{i=1}^{N}$, is given by:
\begin{equation}
f(M \mid \boldsymbol{\mu}, \mathbf{b}) = \prod_{i=1}^{N} \prod_{j=1}^{2} \frac{1}{2b_j^{(i)}} \exp \left( - \frac{\left| p_j^{(i)} - \mu_j^{(i)} \right|}{b_j^{(i)}} \right),
\end{equation}
where $p_j^{(i)} \in \mathbb{R}$ represents the ground-truth location of the map element for the $j^{th}$ coordinate dimension of the $i^{th}$ point. The parameters $\mu_j^{(i)} \in \mathbb{R}$ and $b_j^{(i)} \in \mathbb{R}$ denote the predicted location and scale of the Laplace distribution for the same coordinate dimension, respectively.

\textbf{Injecting Map Uncertainty into Perception.}
As shown in Fig. \ref{mainfig}, we use a set of map queries $Q_m$ to extract map information from $F_{scene}$ and employ an MLP decoder to produce predicted map vectors $V_m \in \mathbb{R}^{N_v \times N_p \times 4}$. 
Here, $N_v = 100$, $N_p = 20$, and 4 represent the number of predicted map vectors, the number of points in each vector, and the four-dimensional vector ($\mu_1, \mu_2, b_1, b_2$ ) for each point, respectively. 
During model training, we replace the L1 loss function with a Negative Log-Likelihood (NLL) loss based on the Laplace distribution.
\begin{equation}
L_R(M \mid \boldsymbol{\mu}, \mathbf{b}) = \sum_{i=1}^{N} \sum_{j=1}^{2} \log \left( 2b_j^{(i)} \right) + \frac{\left| p_j^{(i)} - \mu_j^{(i)} \right|}{b_j^{(i)}}.
\end{equation}
After obtaining the coordinates $\boldsymbol{\mu} = (\mu_1, \mu_2)$ of the map element points and their associated uncertainties $\mathbf{b} = (b_1, b_2)$, we use an MLP encoder to encode these uncertainties into a set of uncertainty queries $Q_{\text{unc}}$ to explicitly capture the map's uncertainty features. 
These queries are then concatenated with the map query $Q_m$, generating an updated map query $Q_m'$ that incorporates the uncertainty information.
\begin{equation}
\begin{aligned}
     &Q_{unc} = \textbf{MLPEncoder}(\mathbf{b}),\\
    &Q_m' = \textbf{MLP}([Q_m, Q_{unc}]),    
\end{aligned}
\end{equation}
where $[\cdot,\cdot]$ denotes the concatenation operation.
\subsection{Uncertainty-Guided Prediction and Planning}\label{3C}
\textbf{Uncertainty-Guided Prediction.} 
Motivated by VAD\cite{jiang2023vad}, we use a set of agent queries $Q_a$ to learn agent-level features from the intermediate representation $F_{scene}$ and map queries $Q_m'$.
To enrich the features required for motion prediction, we first perform feature interaction among agents using an attention mechanism, resulting in updated agent queries $Q_a'$. 
Next, $Q_a'$ interacts with the updated map queries $Q_m'$ to integrate agent-map interactions, leading to further refined agent queries $Q_a''$. Finally, a MLP decoder outputs the multi-modal trajectories $\tau_a \in \mathbb{R}^{N_a \times N_m \times T_f \times 2}$ and confidence scores for each modality, where $N_a$, $N_m$ and $T_f$ represent the number of predicted agents, the number of modalities, and the number of future timestamps, respectively.
The formal expression of the above process is given by:
\begin{equation}
\begin{aligned}
    &Q_a' = \textbf{Self-Attention}(Q_a),\\
   & Q_a'' = \textbf{Cross-Attention}(q=Q_a', k=Q_m', v=Q_m'),\\
    &\tau_a = \textbf{MLP}(Q_a'').
\end{aligned}
\end{equation}

\textbf{Uncertainty-Guided Planning.} 
Consistent with previous works, we do not explicitly introduce ego status information, and derive ego queries $Q_e$ by encoding the ego vehicle’s historical trajectory. 
Following SparseDrive \cite{sun2024sparsedrive}, multi-modal trajectory clustering from the nuScenes dataset is encoded into mode queries $Q_{\text{mode}}$.
To capture multi-modal features, we first fuse $Q_e$ with $Q_{\text{mode}}$, followed by self-attention-based interactions to generate updated queries $Q_e'$. 
Next, attention mechanisms facilitate interactions between $Q_e'$ and $Q_a''$, resulting in $Q_e''$, and further interactions with $Q_m'$ to produce $Q_e'''$. 
This process leverages map uncertainty to guide planning. 
Finally, a MLP decoder outputs multi-modal planning trajectories $\tau_e \in \mathbb{R}^{3 \times N_e \times T_f \times 2}$ with confidence scores for each modality. 
Similar to most autonomous driving systems, we use three driving commands (left turn, right turn, and go straight) as navigation information. 
Hence, 3, $N_e$, and $T_f$ denote the number of driving commands, the number of planning modalities for each command, and the number of future timestamps, respectively.
The formula representing the described process is:
\begin{equation}
\begin{aligned}
   & Q_e' = \textbf{Self-Attention}([Q_e, Q_{mode}]),\\
   & Q_e'' = \textbf{Cross-Attention}(q=Q_e', k=Q_a'', k=Q_a''),\\
   & Q_e''' = \textbf{Cross-Attention}(q=Q_e'', k=Q_m', v=Q_m'),\\
   & \tau_e = \textbf{MLP}(Q_e''').
\end{aligned}
\end{equation}

\subsection{Uncertainty-Collision-Aware Selection}\label{3D}
As shown in Fig. \ref{mainfig}, after generating multi-modal trajectories and assigning the corresponding scores, they are evaluated based on driving area uncertainty and collision risk.
The trajectory with the lowest area uncertainty and collision risk is selected as the safest final output.
Following prior works\cite{chen2024vadv2,hu2023planning,zheng2024genad,jiang2023vad,hu2022st,sun2024sparsedrive}, we first select a subset of multi-modal trajectories for the UCAS strategy, denoted as $\tau_{a, cmd} \in \mathbb{R}^{N_e \times T_f \times 2}$, based on high-level control commands.

\textbf{Uncertainty-Aware Selection.} 
As shown in Fig. \ref{vis_fig1}, to account for uncertainty in the online map, each map boundary point is converted into an elliptical region defined by a Laplace distribution. Driving in areas with high uncertainty is dangerous due to potential map errors, which could lead to a collision.
Driving risk in uncertain areas is assessed by calculating the Negative Log-Likelihood (NLL) values for a set of trajectory points $N$ relative to these elliptical regions. A lower NLL value indicates a higher driving risk, therefore we reduce the score of this trajectory with NLL values below a certain threshold. In practice, we simply set the score of trajectory with a high driving risk to 0.
The NLL computation method is similar to the formula mentioned in Section \ref{3B}.


\textbf{Collision-Aware Selection.} 
We use the agent collision checking module, following SparseDrive \cite{sun2024sparsedrive}, to assess collisions with agents, and the L2 distance with map boundaries to evaluate a potential collisions with map boundaries. If a trajectory poses a high collision risk with predicted agents or map boundaries, its confidence score is set to zero.

Through the UCAS strategy, we select the trajectory with the highest score as the safest trajectory output.

\subsection{Training Loss}\label{3E}
Following VAD\cite{jiang2023vad}, our overall loss function is defined as a weighted sum of the scene learning loss $L_{\text{map}}$, motion prediction learning loss $L_{\text{mot}}$, and planning loss $L_{\text{plan}}$. The total loss is expressed as:
\begin{equation}
L = \lambda_1 L_{\text{map}} + \lambda_2 L_{\text{mot}} + \lambda_3 L_{\text{plan}},
\end{equation}
where $\lambda_1$, $\lambda_2$, and $\lambda_3$ are the weight coefficients for the respective losses. 
It is worth noting that for $L_{map}$, we use the NLL loss, as discussed in Section\ref{3B}, instead of the L1 loss for scene learning.

\begin{table*}[htb]
    \centering
    \setlength{\tabcolsep}{6pt} 
    \renewcommand{\arraystretch}{1.2} 
    \caption{end-to-end planning results on nuScenes dataset\cite{caesar2020nuscenes}}
    \resizebox{0.99\textwidth}{!}{ 
    \begin{tabular}{l|ccc>{\columncolor[gray]{0.9}}c|ccc>{\columncolor[gray]{0.9}}c|ccc>{\columncolor[gray]{0.9}}c}
        \toprule
        \multirow{2}{*}{\text{Method}} & \multicolumn{4}{c|}{DE (m) $\downarrow$} & \multicolumn{4}{c|}{\text{CR (\%) $\downarrow$}} & \multicolumn{4}{c}{\text{DACR (\%) $\downarrow$}} \\
        & \text{1s} & \text{2s} & \text{3s} & \text{Avg.} & \text{1s} & \text{2s} & \text{3s} & \text{Avg.} & \text{1s} & \text{2s} & \text{3s} & \text{Avg.} \\
        \midrule
        NMP$^{*}$ \cite{zeng2019end} & - & - & 2.31 & - & - & - & 1.92 & - & - & - & - & - \\
        SA-NMP$^{*}$ \cite{zeng2019end} & - & - & 2.05 & - & - & - & 1.59 & - & - & - &  - & - \\
        FF$^{*}$ \cite{hu2021safe} & 0.55 & 1.20 & 2.54 & 1.43 & 0.06 & 0.17 & 1.07 & 0.43 & - & - & - & - \\
        EO$^{*}$ \cite{khurana2022differentiable} & 0.67 & 1.36 & 2.78 & 1.60 & 0.04 & 0.09 & 0.88 & 0.33 & - & - & - & - \\
        \midrule
        ST-P3 \cite{hu2022st} & 1.33 & 2.11 & 2.90 & 2.11 & 0.23 & 0.62 & 1.27 & 0.71 & - & - & - & - \\
        UniAD$^{\dagger}$ \cite{hu2023planning} & 0.48 & 0.96 & 1.65 & 1.03 & \textbf{0.05} & 0.17 & 0.71 & 0.31 & 0.60 & 1.75 & 2.44 & 1.60 \\
        VAD$^{\dagger}$ \cite{jiang2023vad} & 0.41 & 0.70 & 1.05 & 0.72 & 0.07 & 0.18 & 0.43 & 0.23 & 0.48 & 1.36 & 2.32 & 1.39 \\
        \textbf{Ours}$^{\dagger}$ & \textbf{0.33} & \textbf{0.59} & \textbf{0.94} & \textbf{0.62} & 0.10 & \textbf{0.14} & \textbf{0.28} & \textbf{0.17} & \textbf{0.23} & \textbf{0.75} & \textbf{1.43} & \textbf{0.80} \\
        \midrule
        SparseDrive$^{\ddagger}$ \cite{sun2024sparsedrive} & 0.30 & 0.58 & \textbf{0.95} & 0.61 & 0.01 & 0.05 & 0.23 & 0.10 & 0.25 & 0.81 & 2.13 & 1.06 \\
        \textbf{Ours}$^{\ddagger}$ & \textbf{0.29} & \textbf{0.57} & \textbf{0.95} & \textbf{0.60} & \textbf{0.00} & \textbf{0.04} & \textbf{0.18} & \textbf{0.07} & \textbf{0.20} & \textbf{0.62} & \textbf{1.84} & \textbf{0.88} \\
        \bottomrule
    \end{tabular}
    }
    \label{tab1}
    \vspace{-1mm}
    \begin{flushleft}
    \footnotesize{
        $*$ Represents the LiDAR-based methods. \\ 
        $\dagger$ Represents using the same metric calculation method as UniAD. \\
        $\ddagger$ Represents using the same metric calculation method as SparseDrive. \\
    }
    \end{flushleft}
\end{table*}

\section{Experiments}

\subsection{Experiments Setup}\label{4A}
\textbf{Benchmark.} We conduct experiments on the challenging nuScenes dataset \cite{caesar2020nuscenes}, which contains 1,000 driving scenes. 

\textbf{Metrics.} Consistent with previous work \cite{hu2023planning,jiang2023vad}, we evaluate the performance through Displacement Error (DE) and Collision Rate (CR). 
Additionally, in order to further assess the quality of trajectories, we introduce a new metric \textit{drivable area conflict rate} (DACR), which considers collision rate between trajectories and map boundary. 
Specifically, DACR checks whether any corner point $\mathbf{c} = (c_x, c_y)$ of the vehicle $\mathbf{C} = \left\{ \mathbf{c}^{(i)} \right\}_{i=1}^{4}$ exceeds the drivable area ($\mathbf{DA}$) during traversal.
DACR in one frame is formulated as:
\begin{equation}
    \delta = 
\begin{cases} 
0, & \forall \mathbf{c} \in \mathbf{C}, \mathbf{c} \in \mathbf{DA}\\
1, & \text{otherwise},
\end{cases}
\end{equation}
\begin{equation}
    DACR = \frac{\sum_{t=0}^{T_f-1} \delta_{t}(\mathbf{C})}{T_f},
\end{equation}
where DA denotes a MultiPolygon type, representing the drivable area.

\textbf{Implementation Details.} All experiments of UncAD are conducted with 8 NVIDIA RTX 3090 GPUs.
More specific parameters are provided in the code release.

\subsection{Main Results}
In this section, we compare our method with competitive baselines on nuScenes dataset in Table \ref{tab1}
Incorporating our method significantly enhances the planning performance of both VAD and SparseDrive, establishing new state-of-the-art (SOTA) results. 

Our Method Performs Better in Planning Safety and Rationality.
The most critical improvements are observed in the collision rate and drivable area conflict rate—key indicators of autonomous driving safety and trajectory rationality. 
VAD+UncAD achieves a \textbf{26\%} reduction in collision rate (0.23\% → 0.17\%) and a \textbf{42\%} reduction in drivable area conflict rate (1.39\% → 0.80\%), 
alongside a \textbf{14\%} improvement in the average planning displacement error (0.72m → 0.62m),
indicating that producing safer and more reliable trajectories. 
For SparseDrive+UncAD, 
while the reduction in average displacement error is modest at \textbf{2\%} (0.61m → 0.60m), 
the collision rate drops by a significant \textbf{30\%} (0.10\% → 0.07\%), and the drivable area conflict rate reduces by \textbf{17\%} (1.06\% → 0.88\%), further demonstrating the robustness of our approach in critical safety metrics.


\begin{figure}
  \centering
  \includegraphics[width=0.45\textwidth]{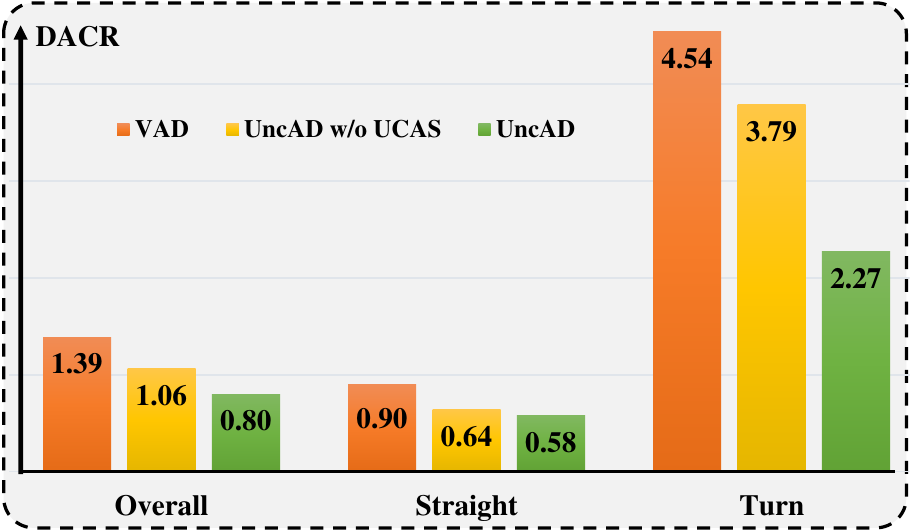}
  \caption{The DACR result of our method and VAD on nuScenes dataset in turn and straight driving scenarios}
  \label{fig:statistics}
\end{figure}
\vspace{-2mm}

\begin{figure*}
  \centering
  \includegraphics[width=0.9\textwidth]{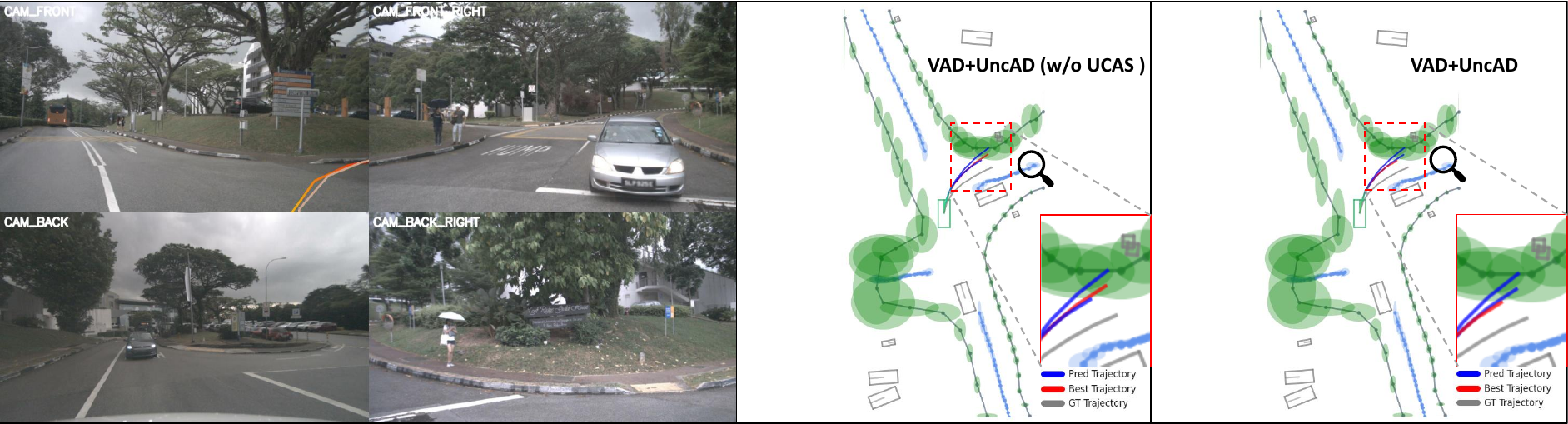}
  \caption{The efficiency of UCAS strategy in avoiding collision.}
  \label{vis_fig1}
\end{figure*}

\begin{figure*}
  \centering
  \includegraphics[width=0.9\textwidth]{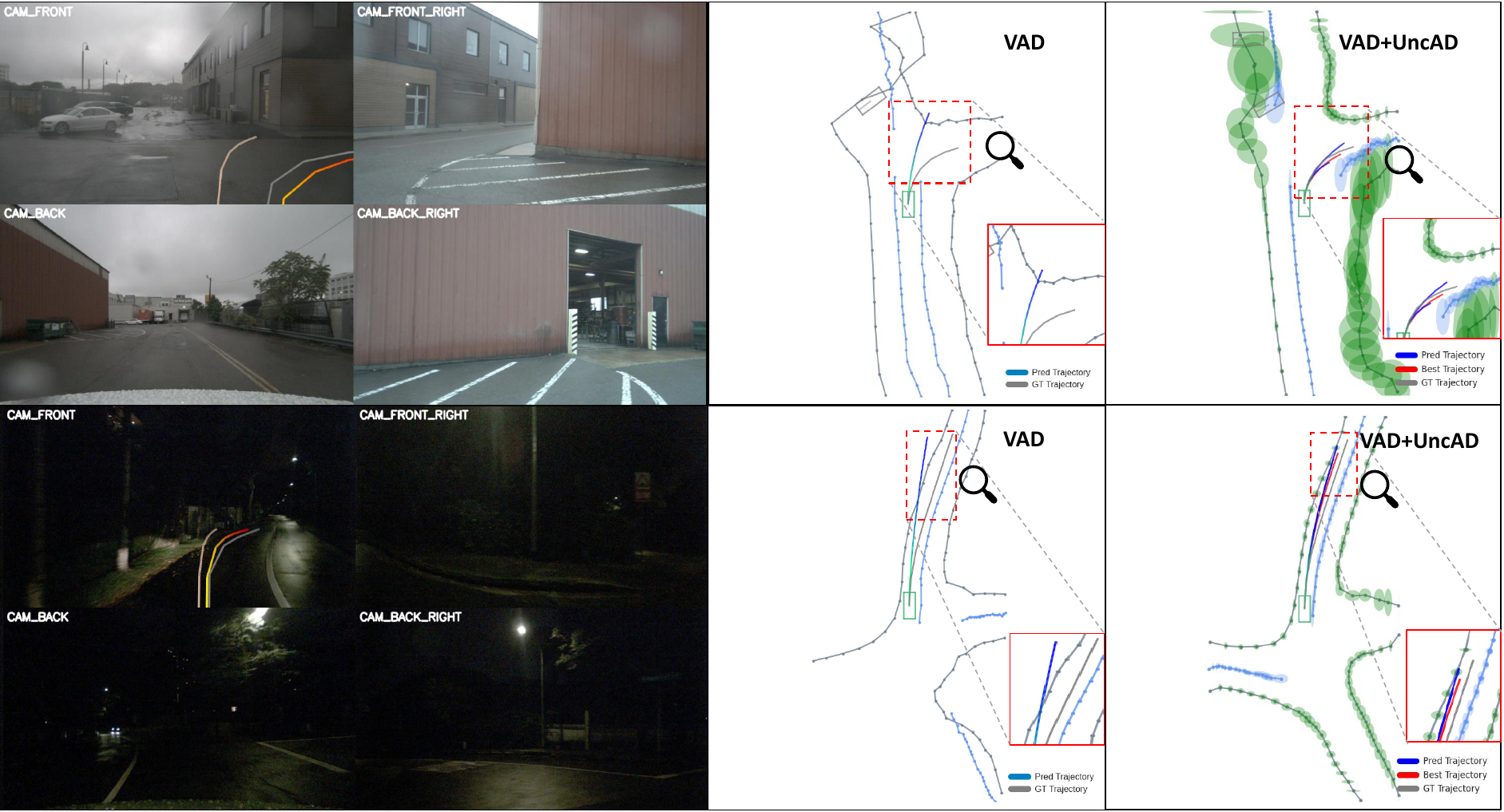}
  \caption{The efficiency of our UncAD method in complex scenarios, such as sharp turn and low texturless night scenarios.}
  \label{vis_fig2}
\end{figure*}


\begin{table}[t]
    \centering
    \setlength{\tabcolsep}{6pt} %
    \renewcommand{\arraystretch}{1.5} %
    \caption{ablation study of each proposed component of UncAD}
    \resizebox{0.5\textwidth}{!}{
    \begin{tabular}{c|cccc|c|c}
        \toprule
        ID & MUE & UGPnP & CAS & UCAS & CR(\%) ↓ & DACR(\%) ↓ \\ 

        \midrule
        1 & & & & & 0.23 & 1.39 \\
        2 & \checkmark & & & & 0.25 & 1.01 \\
        3 & \checkmark & \checkmark & & & 0.19 & 1.06 \\
        4 & \checkmark & \checkmark & \checkmark & & 0.18 & 1.06 \\
        5 & \checkmark & \checkmark & & \checkmark & \textbf{0.17} & \textbf{0.80} \\
        \bottomrule
    \end{tabular}
    }
    \vspace{-4mm}
    \label{tab3}
\end{table}

\subsection{Ablation Result.} 
We provide ablations on nuScenes for the designs of each proposed component.
As shown in Table \ref{tab3}, "MUE" means Map Uncertainty Estimation;"UGPnP" means Uncertainty-Guided Prediction and Planning; "CAS" means collision-aware selection; "UCAS" means uncertainty-collision-aware selection. 
ID-1 represents the planning results of VAD. 

\textbf{The Necessity of Map Uncertainty Estimation Strategy.} ID-2 demonstrates the advantage of incorporating map uncertainty in improving trajectory rationality, reducing the drivable area conflict rate by 27\% compared to ID-1. This improvement is attributed to the ability of map uncertainty to guide the vehicle in better accounting for uncertain regions, resulting in a significant enhancement in drivable area conflict rate performance.

\textbf{The Necessity of Uncertatinty-Guided Prediction and Planning Strategy.} ID-3 of Tab. \ref{tab3} shows the effectiveness of the Uncertatinty-Guided Prediction and Planning module.
Multi-modal planning offers multiple driving modes, allowing the vehicle to select the most appropriate one in complex scenarios, thereby reducing collision rates. 
Map uncertainty plays a crucial guiding role by helping each mode account for uncertain regions, leading to safer and more reliable trajectory planning. This module leverages the map uncertainty to guide Multi-modal planning, results in significant improvements across key metrics: a 15\% reduction in average displacement error, a 17\% decrease in collision rate, and a 24\% reduction in Drivable Area Conflict Rate compared to ID-1.

\textbf{The Necessity of Uncertainty-Collision-Aware Selection Strategy.} 
The Uncertainty-Collision-Aware Selection (UCAS) strategy (ID-5) proves essential and effective in enhancing trajectory safety and reliability. 
UCAS not only detects collision risks but also considers uncertainties in the driving area, enabling the selection of safer and more rational trajectories. 
In contrast, ID-4, which utilizes a collision-aware selection focused solely on surrounding agents, improves the collision rate but does not reduce the drivable area conflict rate compared to ID-3. 
As shown in Fig. \ref{fig:statistics}, the baseline method is \textbf{\textcolor{orange}{VAD}} (ID-1 of Tab. \ref{tab3}) The \textcolor[RGB]{238,188,0}{yellow pillar} indicates UncAD without UCAS strategy (ID-3 of Tab. \ref{tab3}) and the \textcolor[RGB]{115,179,73}{green pillar} indicates the complete UncAD framework (ID-5 of Tab. \ref{tab3}). While UncAD improved performance in both straight and turning scenarios, the inclusion of UCAS led to a substantial 50\% improvement in turning scenarios, significantly reducing conflicts between the ego vehicle and drivable areas. 
This demonstrates the robustness of our approach, particularly in complex real-world situations such as sharp turns(shown in Fig. \ref{vis_fig1}), where safety is critical. Our method not only enhances overall performance but also ensures safer driving in challenging environments.

\subsection{Qualitative Results.}
To illustrate the performance of our method more intuitively, we provide qualitative visualizations of the predicted planning trajectories on an online map with uncertainty in Fig. \ref{vis_fig1} and Fig.  \ref{vis_fig2}.
In the panoramic image(left), the \textbf{\textcolor{pink}{pink trajectory}}  serves as the baseline for comparison. In Fig. \ref{vis_fig1}, it represents the output trajectory without the UCAS strategy, while in Fig. \ref{vis_fig2}, it corresponds to the trajectory generated by VAD. The gray and red trajectories maintain consistent meanings across the BEV maps.
In the BEV maps(right), the \textbf{\textcolor[RGB]{130,190,73}{ego vehicle}} is depicted as a \textcolor[RGB]{130,190,73}{green car}, with \textbf{\textcolor{blue}{predicted multi-modal trajectories}} in \textcolor{blue}{blues}, the \textbf{\textcolor{red}{final predicted trajectory}} in \textcolor{red}{red}, and the \textbf{\textcolor{gray}{ground truth (GT) trajectory}} in \textcolor{gray}{gray}. 
Green ellipses represent the uncertainty of road boundary , and blue ellipses indicate the uncertainty of lane divider.

\textbf{The Efficiency of UCAS Strategy.} In Fig. \ref{vis_fig1}, after incorporating the Uncertainty-Collision-Aware Selection (UCAS) strategy, the ego vehicle selects a safer trajectory, avoiding uncertain areas and effectively preventing potential collisions. In contrast, without the UCAS strategy, the output trajectory is much more likely to collide with map boundaries. This demonstrates the effectiveness of the UCAS strategy in ensuring safer driving.

\textbf{The Efficiency of Our Method in complex scenarios.} 
The above section in the Fig. \ref{vis_fig2} demonstrates that in a sharp turn, guided by uncertainty, the ego vehicle adheres to boundary constraints and tends to stay in safer, more certain areas.
In contrast, VAD generates incorrect planning and leads to collisions. 
The below section in the Fig. \ref{vis_fig2} highlights that our UncAD method performs well even in low-visibility conditions at night, effectively avoiding collisions. This demonstrates the robustness and generalization of our method across various challenging scenarios.

\section{Conclusion}
In this paper, we explore the estimation of map uncertainty in end-to-end autonomous driving, resulting in a novel approach named UncAD. 
Delving into this framework, we effectively utilize the map uncertainty to produce robust and safe planning result via Uncertainty-Guided Planning strategy and Uncertainty-Collision-Aware Planning Selection module. 
Sufficient experiments and SOTA performance on nuScenes have demonstrated the effectiveness of our method.

\balance
\bibliographystyle{IEEEtran}
\bibliography{ref}

\begin{thebibliography}{10}
\providecommand{\url}[1]{#1}
\csname url@samestyle\endcsname
\providecommand{\newblock}{\relax}
\providecommand{\bibinfo}[2]{#2}
\providecommand{\BIBentrySTDinterwordspacing}{\spaceskip=0pt\relax}
\providecommand{\BIBentryALTinterwordstretchfactor}{4}
\providecommand{\BIBentryALTinterwordspacing}{\spaceskip=\fontdimen2\font plus
\BIBentryALTinterwordstretchfactor\fontdimen3\font minus \fontdimen4\font\relax}
\providecommand{\BIBforeignlanguage}[2]{{%
\expandafter\ifx\csname l@#1\endcsname\relax
\typeout{** WARNING: IEEEtran.bst: No hyphenation pattern has been}%
\typeout{** loaded for the language `#1'. Using the pattern for}%
\typeout{** the default language instead.}%
\else
\language=\csname l@#1\endcsname
\fi
#2}}
\providecommand{\BIBdecl}{\relax}
\BIBdecl

\bibitem{chen2024end}
L.~Chen, P.~Wu, K.~Chitta, B.~Jaeger, A.~Geiger, and H.~Li, ``End-to-end autonomous driving: Challenges and frontiers,'' \emph{IEEE Transactions on Pattern Analysis and Machine Intelligence}, vol.~46, no.~12, pp. 10\,164--10\,183, 2024.

\bibitem{chib2023recent}
P.~S. Chib and P.~Singh, ``Recent advancements in end-to-end autonomous driving using deep learning: A survey,'' \emph{IEEE Transactions on Intelligent Vehicles}, vol.~9, no.~1, pp. 103--118, 2023.

\bibitem{zheng2024preliminary}
Y.~Zheng, Z.~Xia, Q.~Zhang, T.~Zhang, B.~Lu, X.~Huo, C.~Han, Y.~Li, M.~Yu, B.~Jin \emph{et~al.}, ``Preliminary investigation into data scaling laws for imitation learning-based end-to-end autonomous driving,'' \emph{arXiv preprint arXiv:2412.02689}, 2024.

\bibitem{jiang2023vad}
B.~Jiang, S.~Chen, Q.~Xu, B.~Liao, J.~Chen, H.~Zhou, Q.~Zhang, W.~Liu, C.~Huang, and X.~Wang, ``Vad: Vectorized scene representation for efficient autonomous driving,'' in \emph{Proceedings of the IEEE/CVF International Conference on Computer Vision (ICCV)}, 2023, pp. 8340--8350.

\bibitem{sun2024sparsedrive}
W.~Sun, X.~Lin, Y.~Shi, C.~Zhang, H.~Wu, and S.~Zheng, ``Sparsedrive: End-to-end autonomous driving via sparse scene representation,'' \emph{arXiv preprint arXiv:2405.19620}, 2024.

\bibitem{zheng2024genad}
W.~Zheng, R.~Song, X.~Guo, and L.~Chen, ``Genad: Generative end-to-end autonomous driving,'' \emph{arXiv preprint arXiv:2402.11502}, 2024.

\bibitem{chen2024vadv2}
S.~Chen, B.~Jiang, H.~Gao, B.~Liao, Q.~Xu, Q.~Zhang, C.~Huang, W.~Liu, and X.~Wang, ``Vadv2: End-to-end vectorized autonomous driving via probabilistic planning,'' \emph{arXiv preprint arXiv:2402.13243}, 2024.

\bibitem{li2023planning}
D.~Li, Q.~Zhang, Z.~Xia, K.~Zhang, M.~Yi, W.~Jin, and D.~Zhao, ``Planning-inspired hierarchical trajectory prediction for autonomous driving,'' \emph{IEEE Transactions on Intelligent Vehicles}, vol.~9, no.~1, pp. 692--703, 2024.

\bibitem{chen2024dualat}
Z.~Chen, Z.~Yu, J.~Li, L.~You, and X.~Tan, ``Dualat: Dual attention transformer for end-to-end autonomous driving,'' in \emph{2024 IEEE International Conference on Robotics and Automation (ICRA)}.\hskip 1em plus 0.5em minus 0.4em\relax IEEE, 2024, pp. 16\,353--16\,359.

\bibitem{ye2023fusionad}
T.~Ye, W.~Jing, C.~Hu, S.~Huang, L.~Gao, F.~Li, J.~Wang, K.~Guo, W.~Xiao, W.~Mao \emph{et~al.}, ``Fusionad: Multi-modality fusion for prediction and planning tasks of autonomous driving,'' \emph{arXiv preprint arXiv:2308.01006}, 2023.

\bibitem{hu2022st}
S.~Hu, L.~Chen, P.~Wu, H.~Li, J.~Yan, and D.~Tao, ``St-p3: End-to-end vision-based autonomous driving via spatial-temporal feature learning,'' in \emph{Proceedings of the European Conference on Computer Vision (ECCV)}, 2022, pp. 533--549.

\bibitem{hu2023planning}
Y.~Hu, J.~Yang, L.~Chen, K.~Li, C.~Sima, X.~Zhu, S.~Chai, S.~Du, T.~Lin, W.~Wang \emph{et~al.}, ``Planning-oriented autonomous driving,'' in \emph{Proceedings of the IEEE/CVF Conference on Computer Vision and Pattern Recognition (CVPR)}, 2023, pp. 17\,853--17\,862.

\bibitem{caesar2020nuscenes}
H.~Caesar, V.~Bankiti, A.~H. Lang, S.~Vora, V.~E. Liong, Q.~Xu, A.~Krishnan, Y.~Pan, G.~Baldan, and O.~Beijbom, ``nuscenes: A multimodal dataset for autonomous driving,'' in \emph{Proceedings of the IEEE/CVF Conference on Computer Vision and Pattern Recognition (CVPR)}, 2020, pp. 11\,621--11\,631.

\bibitem{zhang2022mmfn}
Q.~Zhang, M.~Tang, R.~Geng, F.~Chen, R.~Xin, and L.~Wang, ``Mmfn: Multi-modal-fusion-net for end-to-end driving,'' in \emph{2022 IEEE/RSJ International Conference on Intelligent Robots and Systems (IROS)}.\hskip 1em plus 0.5em minus 0.4em\relax IEEE, 2022, pp. 8638--8643.

\bibitem{li2023conditional}
D.~Li, Q.~Zhang, S.~Lu, Y.~Pan, and D.~Zhao, ``Conditional goal-oriented trajectory prediction for interacting vehicles,'' \emph{IEEE Transactions on Neural Networks and Learning Systems}, vol.~35, no.~12, pp. 18\,758--18\,770, 2024.

\bibitem{fu2023interactionnet}
J.~Fu, Y.~Shen, Z.~Jian, S.~Chen, J.~Xin, and N.~Zheng, ``Interactionnet: Joint planning and prediction for autonomous driving with transformers,'' in \emph{2023 IEEE/RSJ International Conference on Intelligent Robots and Systems (IROS)}.\hskip 1em plus 0.5em minus 0.4em\relax IEEE, 2023, pp. 9332--9339.

\bibitem{xu2024drivegpt4}
Z.~Xu, Y.~Zhang, E.~Xie, Z.~Zhao, Y.~Guo, K.-Y.~K. Wong, Z.~Li, and H.~Zhao, ``Drivegpt4: Interpretable end-to-end autonomous driving via large language model,'' \emph{IEEE Robotics and Automation Letters}, 2024.

\bibitem{jin2024tod3cap}
B.~Jin, Y.~Zheng, P.~Li, W.~Li, Y.~Zheng, S.~Hu, X.~Liu, J.~Zhu, Z.~Yan, H.~Sun \emph{et~al.}, ``Tod3cap: Towards 3d dense captioning in outdoor scenes,'' in \emph{Proceedings of the European Conference on Computer Vision (ECCV)}.\hskip 1em plus 0.5em minus 0.4em\relax Springer, 2024, pp. 367--384.

\bibitem{jin2023adapt}
B.~Jin, X.~Liu, Y.~Zheng, P.~Li, H.~Zhao, T.~Zhang, Y.~Zheng, G.~Zhou, and J.~Liu, ``Adapt: Action-aware driving caption transformer,'' in \emph{2023 IEEE International Conference on Robotics and Automation (ICRA)}.\hskip 1em plus 0.5em minus 0.4em\relax IEEE, 2023, pp. 7554--7561.

\bibitem{liu2024deep}
Y.~Liu, Q.~Zhang, Y.~Gao, and D.~Zhao, ``Deep reinforcement learning-based driving policy at intersections utilizing lane graph networks,'' \emph{IEEE Transactions on Cognitive and Developmental Systems}, vol.~16, no.~5, pp. 1759--1774, 2024.

\bibitem{wang2020dynamic}
J.~Wang, Q.~Zhang, D.~Zhao, M.~Zhao, and J.~Hao, ``Dynamic horizon value estimation for model-based reinforcement learning,'' \emph{IEEE Transactions on Neural Networks and Learning Systems}, vol.~35, no.~7, pp. 8812--8825, 2024.

\bibitem{wang2024prototypical}
J.~Wang, Q.~Zhang, Y.~Mu, D.~Li, D.~Zhao, Y.~Zhuang, P.~Luo, B.~Wang, and J.~Hao, ``Prototypical context-aware dynamics for generalization in visual control with model-based reinforcement learning,'' \emph{IEEE Transactions on Industrial Informatics}, vol.~29, no.~9, pp. 10\,717--10\,727, 2024.

\bibitem{kendall2017uncertainties}
A.~Kendall and Y.~Gal, ``What uncertainties do we need in bayesian deep learning for computer vision?'' \emph{Proceedings of the Advances in Neural Information Processing Systems (NeurIPS)}, vol.~30, 2017.

\bibitem{bae2021estimating}
G.~Bae, I.~Budvytis, and R.~Cipolla, ``Estimating and exploiting the aleatoric uncertainty in surface normal estimation,'' in \emph{Proceedings of the IEEE/CVF International Conference on Computer Vision (ICCV)}, 2021, pp. 13\,137--13\,146.

\bibitem{long2024adaptive}
X.~Long, Y.~Zheng, Y.~Zheng, B.~Tian, C.~Lin, L.~Liu, H.~Zhao, G.~Zhou, and W.~Wang, ``Adaptive surface normal constraint for geometric estimation from monocular images,'' \emph{IEEE Transactions on Pattern Analysis and Machine Intelligence}, 2024.

\bibitem{eldesokey2020uncertainty}
A.~Eldesokey, M.~Felsberg, K.~Holmquist, and M.~Persson, ``Uncertainty-aware cnns for depth completion: Uncertainty from beginning to end,'' in \emph{Proceedings of the IEEE/CVF Conference on Computer Vision and Pattern Recognition (CVPR)}, 2020, pp. 12\,014--12\,023.

\bibitem{kim2019laf}
S.~Kim, S.~Kim, D.~Min, and K.~Sohn, ``Laf-net: Locally adaptive fusion networks for stereo confidence estimation,'' in \emph{Proceedings of the IEEE/CVF Conference on Computer Vision and Pattern Recognition (CVPR)}, 2019, pp. 205--214.

\bibitem{poggi2020uncertainty}
M.~Poggi, F.~Aleotti, F.~Tosi, and S.~Mattoccia, ``On the uncertainty of self-supervised monocular depth estimation,'' in \emph{Proceedings of the IEEE/CVF Conference on Computer Vision and Pattern Recognition (CVPR)}, 2020, pp. 3227--3237.

\bibitem{zheng2023steps}
Y.~Zheng, C.~Zhong, P.~Li, H.-a. Gao, Y.~Zheng, B.~Jin, L.~Wang, H.~Zhao, G.~Zhou, Q.~Zhang \emph{et~al.}, ``Steps: Joint self-supervised nighttime image enhancement and depth estimation,'' in \emph{2023 IEEE International Conference on Robotics and Automation (ICRA)}.\hskip 1em plus 0.5em minus 0.4em\relax IEEE, 2023, pp. 4916--4923.

\bibitem{michelmore2020uncertainty}
R.~Michelmore, M.~Wicker, L.~Laurenti, L.~Cardelli, Y.~Gal, and M.~Kwiatkowska, ``Uncertainty quantification with statistical guarantees in end-to-end autonomous driving control,'' in \emph{2020 IEEE international conference on robotics and automation (ICRA)}.\hskip 1em plus 0.5em minus 0.4em\relax IEEE, 2020, pp. 7344--7350.

\bibitem{kurniawati2011motion}
H.~Kurniawati, Y.~Du, D.~Hsu, and W.~S. Lee, ``Motion planning under uncertainty for robotic tasks with long time horizons,'' \emph{The International Journal of Robotics Research}, vol.~30, no.~3, pp. 308--323, 2011.

\bibitem{yang2020d3vo}
N.~Yang, L.~v. Stumberg, R.~Wang, and D.~Cremers, ``D3vo: Deep depth, deep pose and deep uncertainty for monocular visual odometry,'' in \emph{Proceedings of the IEEE/CVF Conference on Computer Vision and Pattern Recognition (CVPR)}, 2020, pp. 1281--1292.

\bibitem{pairet2018uncertainty}
{\`E}.~Pairet, J.~D. Hern{\'a}ndez, M.~Lahijanian, and M.~Carreras, ``Uncertainty-based online mapping and motion planning for marine robotics guidance,'' in \emph{2018 IEEE/RSJ International Conference on Intelligent Robots and Systems (IROS)}.\hskip 1em plus 0.5em minus 0.4em\relax IEEE, 2018, pp. 2367--2374.

\bibitem{loquercio2020general}
A.~Loquercio, M.~Segu, and D.~Scaramuzza, ``A general framework for uncertainty estimation in deep learning,'' \emph{IEEE Robotics and Automation Letters}, vol.~5, no.~2, pp. 3153--3160, 2020.

\bibitem{sinha2024real}
R.~Sinha, A.~Elhafsi, C.~Agia, M.~Foutter, E.~Schmerling, and M.~Pavone, ``Real-time anomaly detection and reactive planning with large language models,'' \emph{arXiv preprint arXiv:2407.08735}, 2024.

\bibitem{gu2024producing}
X.~Gu, G.~Song, I.~Gilitschenski, M.~Pavone, and B.~Ivanovic, ``Producing and leveraging online map uncertainty in trajectory prediction,'' in \emph{Proceedings of the IEEE/CVF Conference on Computer Vision and Pattern Recognition (CVPR)}, 2024, pp. 14\,521--14\,530.

\bibitem{liao2022maptr}
B.~Liao, S.~Chen, X.~Wang, T.~Cheng, Q.~Zhang, W.~Liu, and C.~Huang, ``Maptr: Structured modeling and learning for online vectorized hd map construction,'' in \emph{Proceedings of the International Conference on Learning Representations (ICLR)}, 2022.

\bibitem{liao2023maptrv2}
B.~Liao, S.~Chen, Y.~Zhang, B.~Jiang, Q.~Zhang, W.~Liu, C.~Huang, and X.~Wang, ``Maptrv2: An end-to-end framework for online vectorized hd map construction,'' \emph{International Journal of Computer Vision}, pp. 1--23, 2024.

\bibitem{liu2023vectormapnet}
Y.~Liu, T.~Yuan, Y.~Wang, Y.~Wang, and H.~Zhao, ``Vectormapnet: End-to-end vectorized hd map learning,'' in \emph{Proceedings of the International Conference on Machine Learning (ICML)}.\hskip 1em plus 0.5em minus 0.4em\relax PMLR, 2023, pp. 22\,352--22\,369.

\bibitem{zeng2019end}
W.~Zeng, W.~Luo, S.~Suo, A.~Sadat, B.~Yang, S.~Casas, and R.~Urtasun, ``End-to-end interpretable neural motion planner,'' in \emph{Proceedings of the IEEE/CVF Conference on Computer Vision and Pattern Recognition (CVPR)}, 2019, pp. 8660--8669.

\bibitem{hu2021safe}
P.~Hu, A.~Huang, J.~Dolan, D.~Held, and D.~Ramanan, ``Safe local motion planning with self-supervised freespace forecasting,'' in \emph{Proceedings of the IEEE/CVF Conference on Computer Vision and Pattern Recognition (CVPR)}, 2021, pp. 12\,732--12\,741.

\bibitem{khurana2022differentiable}
T.~Khurana, P.~Hu, A.~Dave, J.~Ziglar, D.~Held, and D.~Ramanan, ``Differentiable raycasting for self-supervised occupancy forecasting,'' in \emph{Proceedings of the European Conference on Computer Vision (ECCV)}.\hskip 1em plus 0.5em minus 0.4em\relax Springer, 2022, pp. 353--369.

\end{thebibliography}
\end{document}